\documentclass[runningheads]{llncs}
\usepackage[T1]{fontenc}
\usepackage{graphicx}
\usepackage{cite}
\usepackage{amsmath, amssymb}
\usepackage{amsfonts}
\usepackage{graphicx}
\usepackage{textcomp}

\usepackage{booktabs}
\usepackage{multirow}
\usepackage{pifont}
\usepackage{bm}
\usepackage{marvosym}
\usepackage{amsmath}
\usepackage{amsfonts}
\usepackage{algorithm}
\usepackage[noend]{algpseudocode}

\newcommand{\cmark}{\ding{51}}
\usepackage[table]{xcolor}

\begin{document}
\title{VARMA-Enhanced Transformer for Time Series Forecasting}
%
%


\author{Jiajun Song\inst{1} \and Xiaoou Liu\inst{1}\Letter}

\institute{$^1$Renmin University of China \\ \email{\{jiajun.song, xiaoou.liu\}@ruc.edu.cn} }

\maketitle              
\begin{abstract}
Transformer-based models have significantly advanced time series forecasting. Recent work, like the Cross-Attention-only Time Series transformer (CATS), shows that removing self-attention can make the model more accurate and efficient. However, these streamlined architectures may overlook the fine-grained, local temporal dependencies effectively captured by classical statistical models like Vector AutoRegressive Moving Average model (VARMA). To address this gap, we propose \textit{\textbf{VARMAformer}}, a novel architecture that synergizes the efficiency of a cross-attention-only framework with the principles of classical time series analysis. Our model introduces two key innovations: (1) a dedicated \textbf{VARMA-inspired Feature Extractor (VFE)} that explicitly models autoregressive (AR) and moving-average (MA) patterns at the patch level, and (2) a \textbf{VARMA-Enhanced Attention (VE-atten)} mechanism that employs a temporal gate to make queries more context-aware. By fusing these classical insights into a modern backbone, \textit{VARMAformer} captures both global, long-range dependencies and local, statistical structures. Through extensive experiments on widely-used benchmark datasets, we demonstrate that our model consistently outperforms existing state-of-the-art methods. Our work validates the significant benefit of integrating classical statistical insights into modern deep learning frameworks for time series forecasting.
\end{abstract}

\section{Introduction}

Time series forecasting is a cornerstone of data-driven decision-making, with critical applications spanning finance, energy consumption, weather prediction, and resource management~\cite{bollerslev1986generalized,tso2007predicting, kalnay2003atmospheric, wheelwright1998forecasting}. The ability to accurately predict future values based on historical data enables proactive strategies, risk mitigation, and operational optimization. Consequently, the pursuit of more powerful and robust forecasting models has long been a central focus of the machine learning community, leading to a rich and diverse landscape of methodologies, from classical statistical models like ARMA to modern deep learning architectures.

The advent of the Transformer~\cite{vaswani2017attention} marks a significant paradigm shift in this landscape. Its unparalleled success in natural language processing and computer vision, driven by the powerful self-attention mechanism, inspires a new wave of time series models. Architectures such as Informer~\cite{zhou2021informer}, Autoformer~\cite{wu2021autoformer}, and FEDformer~\cite{zhou2022fedformer} quickly set new state-of-the-art benchmarks, demonstrating the potential of attention mechanisms to capture complex, long-range dependencies in temporal data. These models establish a prevailing belief that deep, attention-based architectures are the definitive path toward superior forecasting performance.

However, this consensus has recently been challenged. Pioneering work by Zeng et al.~\cite{zeng2023transformers} raised a critical question: "Are Transformers effective for time series forecasting?" Their empirical studies revealed that surprisingly simple linear models, when properly configured, could outperform many of these complex Transformer-based counterparts, particularly in long-term forecasting tasks~\cite{liu2022non,ekambaram2023tsmixer}. This finding ignites a crucial debate within the community, suggesting that the inductive biases of standard Transformers might not be well-aligned with the fundamental properties of time series data. The success of these simpler models implied that architectural complexity is not a guaranteed prerequisite for performance and, in some cases, might even be detrimental.

In light of these considerations, research has shifted towards identifying the specific components of the Transformer architecture that hinder its effectiveness. The primary culprit, as argued by Kim et al~\cite{kim2024self}, in their work on the Cross-Attention-only Time Series transformer (CATS), is the self-attention mechanism itself. Self-attention is, by design, permutation-invariant; it treats input tokens as an unordered set, which is fundamentally at odds with the strict temporal ordering inherent in time series. This mismatch can lead to the loss of crucial temporal information, as the model struggles to preserve the sequential context of past and future values. The CATS model offers a radical solution to this problem: it eliminates self-attention entirely, relying solely on a cross-attention decoder. In this framework, learnable queries representing the future horizon directly attend to the input time series (as keys and values), thereby preserving temporal order and achieving a more efficient and effective architecture.

While CATS represents a significant step forward by resolving the core issue of self-attention, its streamlined design presents a new, implicit research question: by removing all but the cross-attention mechanism, have we also discarded the model's ability to capture fine-grained, local temporal dynamics? Classical statistical models like VARMA~\cite{tiao1981modeling,hamilton1994time} excel precisely at modeling these local AR and MA patterns. The minimalist structure of CATS, while powerful at capturing global relationships between the past and future, may lack the specialized components needed to explicitly model these classical dependencies, which are ubiquitous in real-world time series.

To address this gap, we introduce \textit{\textbf{VARMAformer}}, a novel hybrid architecture that synergizes the structural efficiency of a cross-attention-only framework with the proven modeling power of classical VARMA principles. Our model is built upon the foundational concepts of CATS but enhances it in two crucial ways. First, we introduce an explicit \textbf{VARMA-inspired Feature Extractor (VFE)} that operates at the patch level to compute distinct AR and MA features from the input series. These features provide the model with a rich, localized understanding of the data's underlying dynamics. Second, we propose a \textbf{VARMA-enhanced Attention (VE-atten)} mechanism, which incorporates a temporal gating module to make the cross-attention process more context-aware. This allows the model to dynamically adjust its focus based on the global characteristics of the input sequence. By fusing these classical insights into a modern Transformer backbone, \textit{VARMAformer} aims to capture the best of both worlds: the global dependency modeling of cross-attention and the local, statistical rigor of VARMA.

Our main contributions are summarized as follows:
\begin{itemize}
\item We propose \textit{VARMAformer}, a novel forecasting architecture that extends the cross-attention-only paradigm by integrating principles from classical VARMA models, effectively bridging the gap between deep learning and statistical methods.
\item We introduce two specific technical innovations: a dedicated VFE module to capture local patch-level dynamics, and the VE-atten mechanism with a temporal gate for more context-aware forecasting.
\item Through extensive experiments on widely-used benchmark datasets, we demonstrate that \textit{VARMAformer} not only retains the efficiency of cross-attention-only models but also achieves superior forecasting accuracy, validating the benefit of our hybrid approach.
\end{itemize}

\section{Related Work}

Our research is positioned at the intersection of three key areas: Transformer-based architectures for time series, the recent trend towards architectural simplification, and hybrid models that integrate classical statistical methods.

\textbf{Transformer-based Time Series Models.} Following the seminal work of Vaswani et al.~\cite{vaswani2017attention}, the Transformer architecture is quickly adapted for time series forecasting. Early influential models like Informer~\cite{zhou2021informer} addresses the quadratic complexity of self-attention with its ProbSparse attention mechanism, enabling efficient processing of long sequences. Subsequent works introduce novel inductive biases tailored for time series. Autoformer~\cite{wu2021autoformer} proposes a decomposition architecture and an auto-correlation mechanism to discover period-based dependencies. FEDformer~\cite{zhou2022fedformer} further refines this by performing attention in the frequency domain, while Crossformer~\cite{zhang2023crossformer} explores capturing cross-variate dependencies through its two-stage attention process. These models represent a powerful line of research that establishes Transformers as a dominant force in the field, primarily leveraging sophisticated self-attention mechanisms within deep encoder-decoder structures.

\textbf{Rethinking Transformers and Architectural Simplification.} Despite their success, the necessity of these complex architectures has been fundamentally questioned. A pivotal study by Zeng et al.~\cite{zeng2023transformers} demonstrates that simple linear models could outperform many intricate Transformer-based models, suggesting that the inductive biases of standard Transformers might be misaligned with the nature of time series data. This sparks a critical re-evaluation of architectural components. A key insight from this research is that the permutation-invariant property of self-attention is a primary source of temporal information loss. Building on this, PatchTST~\cite{Yuqietal-2023-PatchTST} shows that a simple encoder-only Transformer applied to patched time series could be remarkably effective, outperforming previous models and even linear models. This indicats that effective pre-processing (patching) and a simplified backbone are more critical than complex attention mechanisms. The most direct predecessor to our work, the Cross-Attention-only Time Series transformer (CATS)~\cite{kim2024self}, takes this simplification to its logical conclusion by eliminating the self-attention mechanism entirely. By using a decoder-only architecture where learnable queries directly attend to the input series, CATS successfully preserves temporal order while drastically reducing model complexity and improving performance. Our proposed model, \textit{VARMAformer}, is built directly upon this cross-attention-only foundation, inheriting its structural efficiency and robustness.

\textbf{Hybrid Models Fusing Deep Learning and Statistical Methods.} Many early hybrid approaches involved a two-stage process, where a statistical model would capture the linear patterns in the data, and a neural network (like an MLP or RNN) would then model the non-linear residuals~\cite{zhang2003time,pai2005hybrid}. While effective, these methods often lack end-to-end trainability and can be cumbersome to implement. More recent works have sought deeper integration. For instance, some models might incorporate state-space model principles within a recurrent framework to improve interpretability and handle uncertainty, such as DeepAR~\cite{salinas2020deepar} and DeepState~\cite{rangapuram2018deep}. However, few have attempted to integrate the core principles of VARMA models—AR and MA dependencies—directly into the mechanisms of a modern Transformer. Unlike prior hybrid models, \textit{VARMAformer} does not simply use a VARMA model as a pre-processing step or a parallel component. Instead, it introduces VARMA dynamics as first-class, learnable modules within its architecture: an explicit feature extractor for patch-level dependencies and a temporal gating mechanism within the attention layer. This principled, end-to-end integration distinguishes our work from previous hybridization efforts.

\section{Methodology}

\begin{figure}[t]
 	\centering
 	\includegraphics[width=12cm]{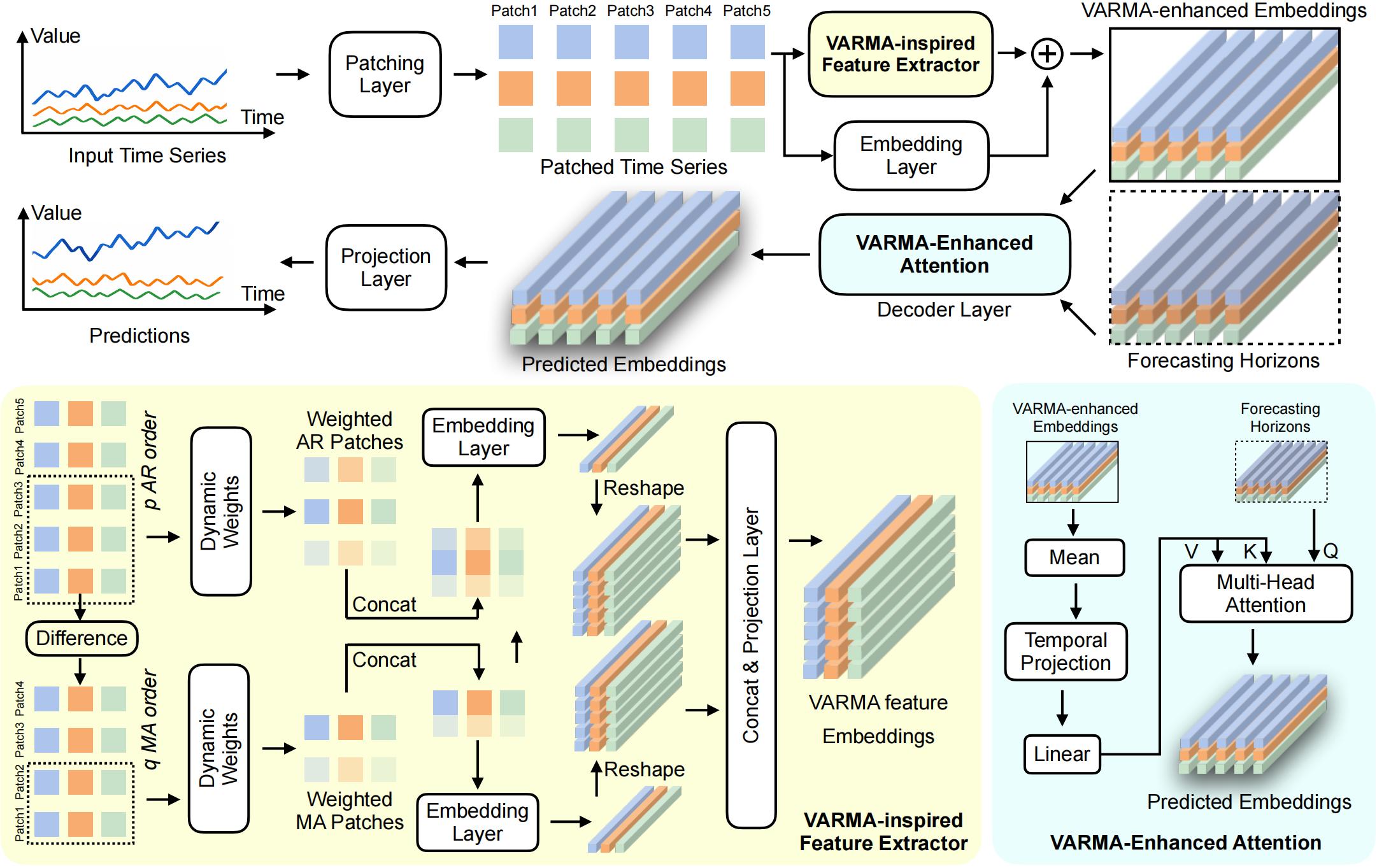}
 	\caption{\textbf{Architecture of proposed \textit{VARMAformer}. First, VARMA-inspired Feature Extraction explicitly models local time series dynamics, and second, the VARMA-Enhanced attention mechanism in the decoder adaptively refines the queries. This principled fusion of statistical priors within an end-to-end attention architecture allows \textit{VARMAformer} to achieve superior performance by leveraging both local and global temporal patterns.} }
 	\label{framework}
\end{figure}

As shown in Fig.~\ref{framework}, we introduce \textit{VARMAformer}, a novel architecture that inherits the foundational principles of CATS while enhancing it with mechanisms inspired by classical VARMA models. Our central hypothesis is that while the cross-attention-only structure excels at capturing global dependencies between the past and future, its performance can be further improved by explicitly modeling the fine-grained, local temporal structures that are the strength of traditional statistical models. To this end, \textit{VARMAformer} introduces two primary innovations: (1) a dedicated VARMA-inspired Feature Extractor that operates on the patch level, and (2) a VARMA-enhanced Attention mechanism that dynamically modulates its queries based on the global context of the input series.

\subsection{Preliminaries and Architecture Framework}

The goal of multivariate time series forecasting is to predict a sequence of future values for multiple correlated variables, given their shared history. Formally, let the historical data be a multivariate time series $\mathbf{X} = \{\mathbf{x}_1, \mathbf{x}_2, \dots, \mathbf{x}_L\} \in \mathbb{R}^{C \times L}$, where $L$ is the length of the look-back window and $C$ is the number of variates (or channels). Each $\mathbf{x}_t \in \mathbb{R}^C$ represents a vector of observations across all $C$ variables at time step $t$.

Given this historical context $\mathbf{X}$, the task is to predict the next $T$ future values, denoted as $\mathbf{Y} = \{\mathbf{y}_{L+1}, \mathbf{y}_{L+2}, \dots, \mathbf{y}_{L+T}\} \in \mathbb{R}^{C \times T}$. A forecasting model $f$ learns a mapping from the input sequence to a prediction $\mathbf{\hat{Y}}$:
\begin{equation}
    f: \mathbf{X} \mapsto \mathbf{\hat{Y}}
\end{equation}
where $\mathbf{\hat{Y}} \in \mathbb{R}^{C \times T}$ is the forecast. The model is trained by minimizing a loss function $\mathcal{L}(\mathbf{Y}, \mathbf{\hat{Y}})$, such as the Mean Squared Error (MSE), between the predicted and the true future values.

\textit{VARMAformer} adopts the core cross-attention-only decoder structure proposed by CATS. We begin by taking a multivariate time series look-back window, $\mathbf{X}$, and applying instance normalization. The normalized series is then partitioned into a sequence of $N$ non-overlapping patches, $\mathbf{X}_p = \{\mathbf{x}_p^{(1)}, \dots, \mathbf{x}_p^{(N)}\}$, where each patch $\mathbf{x}_p^{(i)} \in \mathbb{R}^{C \times P}$ has a length of $P$. To preserve all information, we apply replication padding to the end of the series before patching. This patching strategy transforms the input from a sequence of time points into a sequence of patch-level representations.

Following the CATS framework, our model operates without a self-attending encoder. Instead, the input patch sequence is directly used to form the keys (K) and values (V) for a cross-attention mechanism within the decoder. The queries (Q) are a set of learnable parameters, independent of the input, each corresponding to a specific patch in the future forecast horizon.

\subsection{VARMA-inspired Feature Extraction (VFE)}

A primary contribution of \textit{VARMAformer} is the integration of an explicit feature extraction module that models local AR and MA dynamics at the patch level. 


\subsubsection{AR Feature Extraction}
The AR component is designed to capture the dependency of a current patch on its immediate predecessors. For each patch $\mathbf{x}^{(t)}$ in the input sequence, we extract its AR features by computing a learned, weighted aggregation of the $p$ preceding patches. This process is formulated as:
\begin{equation}
\mathbf{Z}_{\text{AR}}^{(t)} = \text{Proj}_{\text{AR}}\left(\text{Concat}\left(\phi_1 \cdot \mathbf{x}^{(t-1)}, \phi_2 \cdot \mathbf{x}^{(t-2)}, \dots, \phi_p \cdot \mathbf{x}^{(t-p)}\right)\right).
\label{eq:ar_features}
\end{equation}
Here, $p$ is the AR order, and $\{\phi_1, \dots, \phi_p\}$ are learnable scalar weights that capture the importance of each historical lag. The extraction process begins by scaling each historical patch $\mathbf{x}^{(t-i)} \in \mathbb{R}^{C \times P}$ with its corresponding weight $\phi_i$. These $p$ weighted historical patches are then concatenated to form a single feature vector. This concatenated vector, with dimension $p \times P$, is subsequently mapped to the final AR feature representation $\mathbf{Z}_{\text{AR}}^{(t)}$ via a linear projection layer, $\text{Proj}_{\text{AR}}$. The dimensionality transformation of this projection can be described as $\text{Proj}_{\text{AR}}: \mathbb{R}^{p \times P} \to \mathbb{R}^{D_{\text{model}}/2}$. This mechanism allows the model to explicitly learn local, short-term temporal dependencies from the sequence of patches. This mechanism allows the model to explicitly learn local, short-term temporal dependencies from the sequence of patches.

\subsubsection{MA Feature Extraction}
To model the MA component, we require an estimate of past innovation or error terms. We propose a simple yet effective proxy: the error term associated with a past time step is approximated by the first-order difference between consecutive patches, i.e., $\mathbf{\epsilon}^{(t-j)} \approx \mathbf{x}^{(t-j)} - \mathbf{x}^{(t-j-1)}$. The MA feature representation is then constructed by aggregating these historical proxy errors over a window of length $q$:
\begin{equation}
\mathbf{Z}_{\text{MA}}^{(t)} = \text{Proj}_{\text{MA}}\left(\text{Concat}\left(\theta_1 \cdot \mathbf{\epsilon}^{(t-1)}, \theta_2 \cdot \mathbf{\epsilon}^{(t-2)}, \dots, \theta_q \cdot \mathbf{\epsilon}^{(t-q)}\right)\right).
\label{eq:ma_features}
\end{equation}
Here, $q$ is the MA order and $\{\theta_1, \dots, \theta_q\}$ are the corresponding learnable scalar weights. The process begins by scaling each proxy error term $\mathbf{\epsilon}^{(t-j)} \in \mathbb{R}^{P}$ with its weight $\theta_j$. The resulting $q$ weighted vectors are concatenated to form a single vector of dimension $q \times P$. This vector is then mapped by a linear layer, $\text{Proj}_{\text{MA}}$, to the final MA feature representation $\mathbf{Z}_{\text{MA}}^{(t)}$. The dimensionality transformation of this projection is $\text{Proj}_{\text{MA}}: \mathbb{R}^{q \times P} \to \mathbb{R}^{D_{\text{model}}/2}$. This approach enables the model to capture dependencies on the ``shocks'' or unexpected changes in the recent history of the time series.



After separate projections, the resulting AR features $\mathbf{Z}_{\text{AR}} \in \mathbb{R}^{C \times N \times (D/2)}$ and MA features $\mathbf{Z}_{\text{MA}} \in \mathbb{R}^{C \times N \times (D/2)}$ are fused to form a unified representation. This is achieved in two steps. First, the features are concatenated along the last dimension (the feature dimension):
\begin{equation}
\mathbf{Z}_{\text{combined}} = \text{Concat}(\mathbf{Z}_{\text{AR}}, \mathbf{Z}_{\text{MA}})
\end{equation}
where $\mathbf{Z}_{\text{combined}} \in \mathbb{R}^{C \times N \times D}$. This combined tensor is then passed through a final linear fusion layer ($\mathbf{W}_{\text{fuse}}: \mathbb{R}^{C \times N \times D} \to \mathbb{R}^{C \times N \times D}$) to learn the interactions between the AR and MA components, producing the final VARMA enhancement tensor:
\begin{equation}
\mathbf{Z}_{\text{VARMA}} = \mathbf{Z}_{\text{combined}}\mathbf{W}_{\text{fuse}} + \mathbf{b}_{\text{fuse}}
\end{equation}
The resulting tensor $\mathbf{Z}_{\text{VARMA}} \in \mathbb{R}^{C \times N \times D}$ encapsulates both local temporal dependencies and is ready to be integrated with the main patch embeddings.

\subsection{VARMA-Enhanced Transformer Backbone}

The backbone processes the patch sequence to generate forecasts, leveraging the newly extracted VARMA features.

\subsubsection{Embedding and Multi-Source Feature Fusion}

Each input patch $\mathbf{x}_p^{(i)}$ is first linearly projected into a $D$-dimensional embedding space. Crucially, we then fuse this primary representation with the classical features via an additive operation, controlled by a learnable scaling parameter $\alpha$:
\begin{equation}
\mathbf{E}^{(i)} = \mathbf{W}_P(\mathbf{x}_p^{(i)}) + \alpha \cdot \mathbf{Z}_{\text{VARMA}}^{(i)} + \mathbf{PE}^{(i)}
\end{equation}
where $\mathbf{W}_P:\mathbb{R}^P \to \mathbb{R}^D$ is the patch embedding layer and $\mathbf{PE}$ is a learnable positional encoding. $\alpha \in [0,1]$ is a gating parameter that balances the contributions of the two token embeddings. This fusion enriches the patch representations with both local temporal dynamics (from $\mathbf{Z}_{\text{VARMA}}$) and sequence-level order information (from $\mathbf{PE}$). These enriched embeddings, $\mathbf{E}$, form the keys and values for the subsequent attention layers.

\subsubsection{VARMA-Enhanced Decoder}

The decoder is a stack of identical layers. Each layer contains our VARMA-Enhanced Attention mechanism and a standard position-wise Feed-Forward Network (FFN). As in CATS, we employ Layer Normalization and Query-Adaptive Masking—a structured stochastic depth technique that randomly masks the attention output for a given query horizon, forcing the model to rely on its query-specific parameters and preventing overfitting from excessive parameter sharing.



\subsection{VARMA-Enhanced Attention Mechanism (VE-atten)}

Our second key innovation is to directly embed temporal awareness into the attention mechanism. Standard attention treats all parts of the query sequence identically, potentially missing cues from the overall state of the time series. To address this, we introduce a Global Context Gating module that dynamically re-weights the query vectors based on the holistic characteristics of the input sequence.

Specifically, instead of using the queries $\mathbf{Q}$ directly, we first derive a global context vector, $\mathbf{c}_k$, by aggregating the entire key sequence $\mathbf{K}$ along its patch dimension via mean pooling. This context vector, which summarizes the dominant statistical properties of the series, is then transformed into a gate vector $\mathbf{G} \in \mathbb{R}^{D_{\text{model}}}$ using a small two-layer network with a sigmoid activation:
\begin{equation}
\mathbf{G} = \sigma\left( \text{Proj}_{\text{gate}}\left( \overline{\mathbf{K}} \right) \right), \quad \text{where} \quad \overline{\mathbf{K}} = \text{mean}(\mathbf{K})
\label{eq:global_gate}
\end{equation}
Here, $\text{Proj}_{\text{gate}}: \mathbb{R}^{D_{\text{model}}} \to \mathbb{R}^{D_{\text{model}}}$ represents the projection network (e.g., two linear layers), and $\sigma$ is the sigmoid function that maps the output to the range $(0, 1)$.

This global gate $\mathbf{G}$ acts as a data-driven filter, learning to emphasize or suppress certain feature dimensions across all queries. It is applied uniformly to every query vector in the matrix $\mathbf{Q}$ via an element-wise product before the attention scores are computed:
\begin{equation}
\mathbf{Q'} = \beta \cdot \mathbf{Q} \odot \mathbf{G} + (1-\beta) \cdot \mathbf{Q}
\label{eq:gated_query}
\end{equation}
where $\odot$ denotes the Hadamard product and $\beta$ is the weight. This simple yet effective mechanism allows the attention computation to be conditioned on the overall temporal context, enabling the model to focus on more relevant patterns for the given input sequence.

\begin{algorithm}[t] %
\caption{\textit{VARMAformer} Forecasting }
\label{alg:varmaformer_simple}
\begin{algorithmic}[1]
    \State \textbf{Input:} Look-back window $\mathbf{X} \in \mathbb{R}^{C \times L}$.
    \State \textbf{Parameters:} Learnable queries $\mathbf{Q}_{\text{dummy}}$, Model weights $\mathbf{W}$, VARMA weights $\mathbf{\Phi}, \mathbf{\Theta}$.
    \State \textbf{Output:} Forecast $\mathbf{\hat{X}}_{\text{future}} \in \mathbb{R}^{C \times T}$.

    \Statex 
    \Procedure{VARMAformer}{$\mathbf{X}$}
        \Statex \Comment{1. Patching and Normalization}
        \State $\mathbf{X}_{\text{norm}}, \mu, \sigma \gets \text{Normalize}(\mathbf{X})$
        \State $\mathbf{X}_p \gets \text{Patching}(\mathbf{X}_{\text{norm}})$ \Comment{Input series to patches $\mathbf{X}_p$}

        \Statex \Comment{2. VARMA Feature Extraction and Fusion}
        \State $\mathbf{Z}_{\text{VARMA}} \gets \text{VFE}(\mathbf{X}_p, \mathbf{\Phi}, \mathbf{\Theta})$
        \State $\mathbf{E} \gets \text{Linear}(\mathbf{X}_p) + \alpha \cdot \mathbf{Z}_{\text{VARMA}} + \mathbf{PE}$
        \State $\mathbf{K}, \mathbf{V} \gets \mathbf{E}, \mathbf{E}$
        
        \Statex \Comment{3. Cross-Attention with Temporal Gating}
        \State $\mathbf{Q}_{\text{out}} \gets \text{Linear}(\mathbf{Q}_{\text{dummy}})$
        \For{$l = 1 \to N_{\text{layers}}$}
            \State $\mathbf{G}_t \gets \text{sigmoid}(\text{Linear}(\text{mean}(\mathbf{K})))$ \Comment{Create temporal gate}
            \State $\mathbf{Q}_{\text{gated}} \gets \mathbf{Q}_{\text{out}} \odot \mathbf{G}_t$ \Comment{Modulate queries}
            \State $\mathbf{Q}_{\text{out}} \gets \text{DecoderLayer}(\mathbf{Q}_{\text{gated}}, \mathbf{K}, \mathbf{V})$
        \EndFor
        
        \Statex \Comment{4. Final Projection}
        \State $\mathbf{\hat{X}}_{\text{patches}} \gets \text{Linear}(\mathbf{Q}_{\text{out}})$
        \State $\mathbf{\hat{X}}_{\text{future}} \gets \text{DeNormalize}(\text{Flatten}(\mathbf{\hat{X}}_{\text{patches}}), \mu, \sigma)$
        
        \State \textbf{return} $\mathbf{\hat{X}}_{\text{future}}$
    \EndProcedure
\end{algorithmic}
\end{algorithm}

This temporal gating allows the model to refine its ``questions'' (queries) based on the overall nature of the ``information source'' (keys), creating a more sophisticated and context-aware interaction between the past and the future.




\subsection{Overall Algorithm}
\textit{}To provide a holistic view of our proposed method, we consolidate the entire forecasting process of VARMAformer into a single, high-level algorithm. As detailed in Algorithm \ref{alg:varmaformer_simple}, the procedure begins with input normalization and patching. The core of our innovation lies in the subsequent steps, where we first extract classical time series dynamics through the VFE and fuse these features with the patch embeddings. The enhanced representations then serve as queries for the decoder. Within the decoder loop, our novel VE-atten mechanism modulates the VARMA-Enhanced embeddings and the forecasting Horizon, allowing for a more context-aware information retrieval. Finally, the processed queries are projected to generate the forecast, which is then de-normalized to produce the final output.

\section{Experiment}

To rigorously evaluate our proposed model, we perform extensive experiments on seven real-world benchmark datasets. We adhere to the standard evaluation protocol established in prior works~\cite{bai2018empirical,zhou2021informer,wu2021autoformer,zhou2022fedformer} to ensure a fair and direct comparison with existing methods.

\textbf{Experimental setup:} For all models, we use a fixed historical look-back window of 96 time steps. Performance is assessed across a range of long-term forecast horizons, specifically {96, 192, 336, 720} for all datasets.

\textbf{Datasets:} Our evaluation suite includes seven benchmarks: ETT~\cite{zhou2021informer} (Electricity Transformer Temperature, which we split into ETTh1, ETTh2, ETTm1, and ETTm2), Electricity (power consumption), Traffic (road occupancy), and Weather (climatological indicators). 

\textbf{Baselines:} We compare our model against a diverse set of state-of-the-art (SOTA) methods, including CATS~\cite{kim2024self}, PatchTST (Patch.)~\cite{Yuqietal-2023-PatchTST}, Timesnet (Time.)~\cite{wu2023timesnet}, Basisformer (Basis.)~\cite{ni2023basisformer}, Crossformer (Cross.)~\cite{zhang2023crossformer}, FEDformer (FED.)~\cite{zhou2022fedformer}, Autoformer (Auto.)~\cite{wu2021autoformer}, and Informer (In.)~\cite{zhou2021informer}. For all results, we report performance of our model alongside the results of other models as presented in CATS, ensuring a consistent comparison across all baselines. We used 1 80GB NVIDIA H100 Tensor Core GPU for all experiments.

\subsection{Main results}

\begin{table}[t]
\centering
\caption{MSE comparison with SOTA methods. Highest and second best are highlighted in bold and underline, respectively}
\fontsize{6pt}{7.5pt}\selectfont
\setlength{\tabcolsep}{1.8mm}{
\begin{tabular}{c|c|c|c|c|c|c|c|c|c|c}

\multicolumn{2}{c}{\textbf{Models}} & \multicolumn{1}{c}{\textbf{Ours}} & \multicolumn{1}{c}{\textbf{CATS}}  & \multicolumn{1}{c}{\textbf{Patch.}} & \multicolumn{1}{c}{\textbf{Time.}} & \multicolumn{1}{c}{\textbf{Basis.}} & \multicolumn{1}{c}{\textbf{Cross.}} & \multicolumn{1}{c}{\textbf{FED.}} & \multicolumn{1}{c}{\textbf{Auto.}} & \multicolumn{1}{c}{\textbf{In.}} \\ \bottomrule
\multirow{4}{*}{ETTm2} & 96 & \textbf{0.178} & \textbf{0.178} & \underline{0.183} & 0.187 & \textbf{0.178} & 0.287 & 0.203 & 0.255 & 0.365 \\
 & 192 & \textbf{0.244} & 0.249  & 0.255 & 0.249 & \underline{0.247} & 0.414 & 0.269 & 0.281 & 0.533 \\
 & 336 & \textbf{0.301} & \underline{0.304}  & 0.309 & 0.321 & 0.314 & 0.597 & 0.325 & 0.339 & 1.363 \\
 & 720 & \textbf{0.399} & \underline{0.402}  & 0.412 & 0.408 & 0.412 & 1.730 & 0.421 & 0.422 & 3.379 \\ \hline
 
\multirow{4}{*}{ETTm1} & 96 &\textbf{0.314}  & \underline{0.318}  & 0.352 & 0.338 & 0.342 & 0.404 & 0.379 & 0.505 & 0.672 \\
 & 192 & \textbf{0.357} & \textbf{0.357}  & 0.390 & \underline{0.374} & 0.379 & 0.450 & 0.426 & 0.553 & 0.795 \\
 & 336 & \textbf{0.386} & \underline{0.387}  & 0.421 & 0.410 & 0.448 & 0.532 & 0.445 & 0.621 & 1.212 \\
 & 720 & \textbf{0.446} & \underline{0.448}  & 0.462 & 0.478 & 0.503 & 0.666 & 0.543 & 0.671 & 1.166 \\ \hline
 
\multirow{4}{*}{ETTh2} & 96 & \textbf{0.279} & \underline{0.287}  & 0.308 & 0.340 & 0.325 & 0.745 & 0.346 & 0.358 & 3.755 \\
 & 192 & \textbf{0.348} & \underline{0.361}  & 0.393 & 0.402 & 0.391 & 0.877 & 0.429 & 0.456 & 5.602 \\ 
 & 336 & \textbf{0.360} & \underline{0.374}  & 0.427 & 0.452 & 0.423 & 1.043 & 0.496 & 0.482 & 4.721 \\ 
 & 720 & \textbf{0.403} & \underline{0.412}  & 0.436 & 0.462 & 0.445 & 1.104 & 0.463 & 0.515 & 3.647 \\  \hline
 
\multirow{4}{*}{ETTh1} & 96 & \textbf{0.363} & \underline{0.371}  & 0.460 & 0.384 & 0.393 & 0.423 & 0.376 & 0.449 & 0.865 \\
 & 192 & \textbf{0.415} & \underline{0.426}  & 0.512 & 0.436 & 0.442 & 0.471 & 0.420 & 0.500 & 1.008 \\
 & 336 & \textbf{0.429} & \underline{0.437}  & 0.546 & 0.638 & 0.473 & 0.570 & 0.459 & 0.521 & 1.107 \\
 & 720 & \textbf{0.439} & \underline{0.474}  & 0.544 & 0.521 & 0.492 & 0.653 & 0.506 & 0.514 & 1.181 \\ \hline
 
\multirow{4}{*}{Electricity} & 96 & \textbf{0.148} & \underline{0.149}  & 0.190 & 0.168 & 0.165 & 0.219 & 0.193 & 0.201 & 0.274 \\ 
 & 192 & \textbf{0.163} & \textbf{0.163}  & 0.199 & 0.184 & \underline{0.178} & 0.231 & 0.201 & 0.222 & 0.296 \\
 & 336 & \underline{0.181} & \textbf{0.180}  & 0.217 & 0.198 & 0.186 & 0.246 & 0.214 & 0.231 & 0.300 \\
 & 720 & \textbf{0.218} & 0.219  & 0.258 & 0.220 & 0.223 & 0.280 & 0.246 & 0.254 & 0.373 \\ \hline
 
\multirow{4}{*}{Traffic} & 96 & \textbf{0.418} & \underline{0.421}  & 0.526 & 0.593 & 0.444 & 0.644 & 0.587 & 0.613 & 0.719 \\
 & 192 & \textbf{0.433} & \underline{0.436}  & 0.522 & 0.617 & 0.446 & 0.665 & 0.604 & 0.616 & 0.696 \\
 & 336 &\textbf{0.449} & \underline{0.453}  & 0.517 & 0.629 & 0.471 & 0.674 & 0.621 & 0.622 & 0.777 \\
 & 720 & \textbf{0.480} & \underline{0.484}  & 0.552 & 0.640 & 0.486 & 0.683 & 0.626 & 0.660 & 0.864 \\ \hline
 
\multirow{4}{*}{Weather} & 96 & \textbf{0.160} & \underline{0.161}  & 0.186 & 0.172 & 0.173 & 0.195 & 0.217 & 0.266 & 0.300 \\
 & 192 & \underline{0.207} & 0.208  & 0.234 & 0.219 & \textbf{0.181} & 0.209 & 0.276 & 0.307 & 0.598 \\
 & 336 & \underline{0.263} & 0.264  & 0.284 & 0.246 & \textbf{0.205} & 0.273 & 0.339 & 0.359 & 0.578 \\
 & 720 & \textbf{0.340} & 0.342  & 0.356 & 0.365 & 0.355 & 0.379 & 0.403 & 0.419 & 1.059 \\ \bottomrule
\end{tabular}}
\label{sota_MSE}
\end{table}

 \begin{table}[t]
\centering
\caption{MAE comparison with SOTA methods.}
\fontsize{6pt}{7.5pt}\selectfont
\setlength{\tabcolsep}{1.8mm}{
\begin{tabular}{c|c|c|c|c|c|c|c|c|c|c}

\multicolumn{2}{c}{\textbf{Models}} & \multicolumn{1}{c}{\textbf{Ours}} & \multicolumn{1}{c}{\textbf{CATS}}  & \multicolumn{1}{c}{\textbf{Patch.}} & \multicolumn{1}{c}{\textbf{Time.}} & \multicolumn{1}{c}{\textbf{Basis.}} & \multicolumn{1}{c}{\textbf{Cross.}} & \multicolumn{1}{c}{\textbf{FED.}} & \multicolumn{1}{c}{\textbf{Auto.}} & \multicolumn{1}{c}{\textbf{In.}} \\ \bottomrule
\multirow{4}{*}{ETTm2} & 96 & \textbf{0.260} & \underline{0.261}  & 0.270 & 0.267 & 0.262 & 0.366 & 0.287 & 0.339 & 0.453 \\
 & 192 & \textbf{0.303} & 0.308  & 0.314 & 0.309 & \underline{0.307} & 0.492 & 0.328 & 0.340 & 0.563 \\
 & 336 & \textbf{0.341} & \underline{0.343}  & 0.347 & 0.351 & 0.349 & 0.542 & 0.366 & 0.372 & 0.887 \\
 & 720 & \textbf{0.397} & \underline{0.402}  & 0.404 & 0.403 & \underline{0.402} & 1.042 & 0.415 & 0.419 & 1.338 \\ \hline
 
\multirow{4}{*}{ETTm1} & 96 & \textbf{0.352} & \underline{0.357}  & 0.374 & 0.375 & 0.376 & 0.426 & 0.419 & 0.475 & 0.571 \\
 & 192 & \textbf{0.377} & \underline{0.378}  & 0.393 & 0.387 & 0.394 & 0.451 & 0.441 & 0.496 & 0.669 \\
 & 336 & \textbf{0.399} & \underline{0.401}  & 0.414 & 0.411 & 0.429 & 0.515 & 0.459 & 0.537 & 0.871 \\
 & 720 & \textbf{0.436} & \underline{0.437}  & 0.449 & 0.450 & 0.467 & 0.589 & 0.490 & 0.561 & 0.823 \\ \hline
 
\multirow{4}{*}{ETTh2} & 96 & \textbf{0.333} & \underline{0.341}  & 0.355 & 0.374 & 0.362 & 0.584 & 0.388 & 0.397 & 1.525 \\
 & 192 & \textbf{0.319} & \underline{0.388}  & 0.405 & 0.414 & 0.406 & 0.656 & 0.439 & 0.452 & 1.931 \\
 & 336 & \textbf{0.396} & \underline{0.403}  & 0.436 & 0.452 & 0.430 & 0.731 & 0.487 & 0.486 & 1.835 \\
 & 720 & \textbf{0.428} & \textbf{0.433}  & \underline{0.450} & 0.468 & 0.456 & 0.763 & 0.474 & 0.511 & 1.625 \\ \hline
 
\multirow{4}{*}{ETTh1} & 96 & \textbf{0.390} & \underline{0.395}  & 0.447 & 0.402 & 0.411 & 0.448 & 0.419 & 0.459 & 0.713 \\
 & 192 & \textbf{0.419} & \underline{0.422}  & 0.477 & 0.429 & 0.438 & 0.474 & 0.448 & 0.482 & 0.792 \\
 & 336 & \textbf{0.428} & \underline{0.432}  & 0.496 & 0.469 & 0.451 & 0.546 & 0.465 & 0.496 & 0.809 \\
 & 720 & \textbf{0.444} & \underline{0.461}  & 0.517 & 0.500 & 0.481 & 0.621 & 0.507 & 0.512 & 0.865 \\ \hline
 
\multirow{4}{*}{Electricity} & 96 & \textbf{0.237} & \textbf{0.237}  & 0.296 & 0.272 & \underline{0.259} & 0.314 & 0.308 & 0.317 & 0.368 \\
 & 192 & \textbf{0.250} & \textbf{0.250}  & 0.304 & 0.322 & \underline{0.272} & 0.322 & 0.315 & 0.334 & 0.386 \\
 & 336 & \underline{0.269} & \textbf{0.268}  & 0.319 & 0.300 & 0.282 & 0.337 & 0.329 & 0.338 & 0.394 \\
 & 720 & \textbf{0.301} & \underline{0.302}  & 0.352 & 0.320 & 0.311 & 0.363 & 0.355 & 0.361 & 0.439 \\ \hline
 
\multirow{4}{*}{Traffic} & 96 & \textbf{0.266} & \underline{0.270}  & 0.347 & 0.321 & 0.315 & 0.429 & 0.366 & 0.388 & 0.391 \\
 & 192 & \textbf{0.271} & \underline{0.275}  & 0.332 & 0.336 & 0.318 & 0.431 & 0.373 & 0.382 & 0.379 \\
 & 336 & \textbf{0.279} & \underline{0.284}  & 0.334 & 0.336 & 0.317 & 0.420 & 0.383 & 0.387 & 0.420 \\
 & 720 & \textbf{0.299} & \underline{0.303}  & 0.352 & 0.350 & 0.312 & 0.424 & 0.382 & 0.408 & 0.472 \\ \hline
 
\multirow{4}{*}{Weather} & 96 & \textbf{0.206} & \underline{0.207}  & 0.227 & 0.220 & 0.214 & 0.271 & 0.296 & 0.336 & 0.384 \\
 & 192 & \underline{0.248} & 0.250 & 0.265 & 0.261 & \textbf{0.222} & 0.277 & 0.336 & 0.367 & 0.544 \\
 & 336 & \underline{0.289} & 0.290  & 0.301 & 0.337 & \textbf{0.242} & 0.332 & 0.380 & 0.395 & 0.523 \\
 & 720 & \textbf{0.339} & \underline{0.341}  & 0.359 & 0.359 & 0.347 & 0.401 & 0.428 & 0.428 & 0.741 \\ \bottomrule
\end{tabular}}
\label{sota_MAE}
\end{table}

Table~\ref{sota_MSE} and Table~\ref{sota_MAE} presents the comprehensive comparison of our proposed \textit{VARMAformer} against several state-of-the-art (SOTA) time series forecasting models across seven real-world datasets. The evaluation is conducted on four forecasting horizons, ranging from 96 to 720, using Mean Squared Error (MSE) and Mean Absolute Error (MAE) as the primary performance metrics. The results robustly demonstrate the superiority and effectiveness of our proposed architecture.

\textbf{Overall Performance}. Our model achieves SOTA performance, outperforming all baseline models in the vast majority of experimental settings. Across the 56 total test scenarios (7 datasets × 4 prediction lengths × 2 metrics), our model delivered the best results (indicated in bold) in 50 cases.  This comprehensive superiority underscores the powerful predictive capability derived from our novel fusion of VARMA principles and attention mechanisms. When compared to other strong baselines such as PatchTST, and various Transformer variants like Basisformer, Crossformer and FEDformer, our model consistently shows substantial improvements, marked by considerably lower MSE and MAE values across all datasets and horizons.

\textbf{Head-to-Head Comparison with the Strong Baseline CATS.} On the ETTh1 dataset, for instance, our model achieves an MSE of 0.439 for the 720-step forecast, a marked improvement over CATS's 0.474. More notably, our model establishes a dominant performance on the challenging high-frequency datasets (ETTm1, ETTm2, and Weather). On the Weather dataset, our model secures the top rank in all forecasting horizons for both MSE and MAE. This suggests that our VARMA-based feature extractor is highly effective at capturing the local statistical patterns and short-term dynamics inherent in fine-grained data, a domain where purely attention-based models can often struggle.

\textbf{Long-Term Forecasting Capability}. The advantages of our model are not limited to short-term predictions but become particularly pronounced in more challenging long-term forecasting tasks. A clear trend emerges from the results: as the prediction length increases, the performance gap between our model and the baselines widens. For example, on the ETTh1 dataset, the MSE difference between our model (0.363) and CATS (0.371) is modest at the 96-step horizon. However, this gap expands at the 720-step horizon, where our model achieves an MSE of 0.439 compared to 0.474 for CATS. This indicates that the temporal dependencies learned by our model are more robust and generalizable, allowing for more reliable extrapolation into the distant future. This enhanced long-range capability is likely attributable to the principled integration of statistical VARMA priors, which guides the model to learn the underlying data generation process rather than merely fitting the training data.

\textbf{Discussion.} An insightful observation from our results is that our model demonstrates particularly significant performance gains on datasets collected at an hourly frequency (ETTh1, ETTh2) compared to those with minutely granularity (ETTm1, ETTm2, Weather). We attribute this behavior to the interplay between our model's architecture and the intrinsic properties of the data at different time scales. Hourly data, such as in the ETTh datasets, typically exhibits strong and clear periodicities, most notably the 24-hour daily cycle. Our model, with its ability to capture both long-range dependencies via attention and local statistical patterns via the VARMA module, is exceptionally well-suited to learn from these pronounced, high-signal-to-noise patterns. In contrast, minutely data is characterized by a much higher degree of noise and stochastic volatility, which can obscure the underlying long-term cycles. While our model still outperforms baselines on these datasets, the relative margin of improvement is more moderate. This suggests that while our VARMA-enhancements are effective at modeling short-term dynamics, the task of distinguishing true periodic signals from high-frequency noise in fine-grained data remains a fundamental challenge. The model's strength in leveraging clear, structured temporal patterns, which are more readily available in the smoother hourly data, is the primary reason for its more pronounced success on the ETTh datasets. This finding highlights a promising direction for future work: developing more advanced noise-robust mechanisms to further unlock predictive potential in high-frequency time series.

\subsection{Ablation studies}

\begin{table}[t]
\centering
\caption{Ablation study on ETTh1 and ETTh2.}
\fontsize{6pt}{7.5pt}\selectfont
\setlength{\tabcolsep}{1.8mm}{
\begin{tabular}{cccllllllll}
\multicolumn{11}{c}{\color{gray}\textit{Comparison on ETTh1}}\\ \bottomrule
\multicolumn{3}{l}{\textbf{Module}} & \multicolumn{2}{c}{\textbf{96}} & \multicolumn{2}{c}{\textbf{192}} & \multicolumn{2}{c}{\textbf{336}} & \multicolumn{2}{c}{\textbf{720}} \\ 
AR & MA & VE-atten & MSE & MAE & MSE & MAE & MSE & MAE & MSE & MAE \\ \hline
&  &  &0.371  & 0.395 &0.426  &0.422  &0.437  &0.432  &0.474  &0.461   \\
\cmark&  &  &0.364  & \textbf{0.391} &0.417  &0.420  &0.434  &0.430  &0.449  &0.447   \\
 & \cmark &  &0.364  &0.392  &0.417  &0.420  &0.434  &0.430  &0.448  &0.447  \\
\cmark & \cmark & & 0.364 &0.392  &0.416  &0.420  &0.432  &0.430  &0.444  & 0.447   \\
 &  & \cmark & 0.365 & \textbf{0.391} & 0.418 & 0.420 & 0.434 & 0.430 & 0.450 & 0.448 \\
\rowcolor{gray!15}\cmark & \cmark & \cmark &\textbf{0.363}&\textbf{0.391}&\textbf{0.415}&\textbf{0.419}&\textbf{0.429}&\textbf{0.428}&\textbf{0.439} &\textbf{0.446} \\ \bottomrule

\multicolumn{11}{c}{\color{gray}\textit{Comparison on ETTh2}}\\ \bottomrule
&  &  & 0.287  & 0.341 &0.361  &0.388  &0.374  &0.403  &0.412  &0.433  \\ 
\cmark &  &  &0.280& 0.336 & 0.353 & 0.385 & 0.365 & 0.400 & 0.409 & 0.433   \\
 & \cmark &  & 0.280 & 0.334 & 0.351 & 0.383 & 0.364 & 0.399 & 0.406 & 0.430 \\
\cmark & \cmark & & 0.280 & 0.336 & 0.348 & 0.380 & 0.361 & 0.397 & 0.406 & 0.430   \\
 &  & \cmark & 0.283 & 0.339 & 0.350 & 0.381 & 0.367 & 0.402 & 0.408 & 0.432 \\
\rowcolor{gray!15}\cmark & \cmark & \cmark &\textbf{0.279}&\textbf{0.333}&\textbf{0.348}&\textbf{0.379}&\textbf{0.360}&\textbf{0.396}&\textbf{0.403}&\textbf{0.428} \\ \bottomrule

\end{tabular}}
\label{ablation}
\end{table}

To rigorously validate the effectiveness of each proposed component within our \textit{VARMAformer} architecture, we conduct a comprehensive ablation study on the ETTh1 and ETTh2 datasets. We systematically evaluate the performance of our model by selectively deactivating our key innovations: the AR feature extractor, the MA feature extractor, and the VE-atten mechanism. The results, presented in Table~\ref{ablation}, unequivocally demonstrate that each component contributes positively to the model's overall performance, and their combination yields a significant synergistic effect. The study first reveals the individual impact of each module when added to a baseline CATS. On the ETTh1 dataset for the 96-step forecast, introducing the AR module alone reduces the MSE from a baseline of 0.371 to 0.364. Similarly, incorporating only the VE-atten module lowers the MSE to 0.365. This confirms that both the explicit modeling of local AR patterns and the context-aware attention gating are independently beneficial. While the MA module alone shows a similar MSE of 0.364, its primary strength emerges when combined with the AR component, as discussed next. The results clearly indicate a powerful synergy between the proposed modules. On ETTh2, combining the AR and MA modules (AR + MA) reduces the MSE to 0.280 for the 96-step forecast, which is a notable improvement over using either component in isolation (0.280 for AR-only, 0.402 for MA-only). This suggests that jointly modeling dependencies on both past values (AR) and past innovations (MA) provides a more complete statistical representation of the local time series dynamics. The full \textit{VARMAformer} model, which integrates all three components (AR + MA + VE-atten), consistently achieves the best performance across all datasets and forecasting horizons. On ETTh1, the full model achieves an MSE of 0.363 for the 96-step forecast, outperforming all other variants. This trend is even more pronounced in long-term forecasting scenarios; for the 720-step forecast on ETTh2, the full model's MSE of 0.403 is substantially lower than the baseline (0.412) and any partial combination. This culminating result powerfully validates our central thesis: the principled, end-to-end fusion of the VFE and VE-atten creates a more robust and accurate time series forecasting model.

\subsection{Impact of Hyper-parameters}

\subsubsection{Impact of different orders}

To investigate the sensitivity of our model to the orders of the VARMA feature extractor, we conducted experiments with varying AR order $p$ and MA order $q$ on the ETTh1 dataset. As shown in Table~\ref{order}, the model's performance is robust to changes in these hyperparameters, with all tested configurations yielding highly competitive results. We observe that increasing the orders from (p=1, q=1) to (p=2, q=2) provides a consistent, albeit modest, improvement across all forecasting horizons. For instance, in the most challenging 720-step forecast, the (2, 2) configuration achieves the best MSE of 0.439 and MAE of 0.444. This suggests that while a first-order model is sufficient to capture the most significant local dependencies, a second-order model can extract additional, useful information from the immediate preceding patches. However, given the marginal nature of these gains, it also indicates that our model is not overly dependent on high-order statistical modeling, relying on the attention mechanism for longer-term dependencies. Therefore, we selected (p=2, q=2) as a balanced and effective default setting for our main experiments.

\begin{table}[t]
\centering
\caption{Impact of AR order $p$ and MA order $q$ on ETTh1.}
\fontsize{6pt}{7.5pt}\selectfont
\setlength{\tabcolsep}{1.8mm}{
\begin{tabular}{ccllllllll}

\multicolumn{2}{l}{\textbf{Order}} & \multicolumn{2}{c}{\textbf{96}} & \multicolumn{2}{c}{\textbf{192}} & \multicolumn{2}{c}{\textbf{336}} & \multicolumn{2}{c}{\textbf{720}} \\ \bottomrule
AR & MA & MSE & MAE & MSE & MAE & MSE & MAE & MSE & MAE \\ \hline
1 & 1 & \textbf{0.363} & 0.391 & 0.416 & \textbf{0.419} & 0.432 & 0.429 & 0.444 & 0.445 \\
1 & 2 & 0.364 & 0.391 & 0.417 & \textbf{0.419}& 0.432 & 0.429    & 0.446 & 0.446 \\
2 & 1 & \textbf{0.363} & 0.391 & 0.416 & \textbf{0.419} & 0.431 & 0.429 & 0.446 & 0.447 \\
\rowcolor{gray!15}2 & 2 &\textbf{0.363}&\textbf{0.390}&\textbf{0.415}&\textbf{0.419}&\textbf{0.429}&\textbf{0.428}&\textbf{0.439} &\textbf{0.444} \\ \bottomrule
\end{tabular}}
\label{order}
\end{table}

\subsubsection{Impact of scaling parameter $\alpha$ and gate $\beta$}
\begin{figure}[t]
 	\centering
 	\includegraphics[width=12cm]{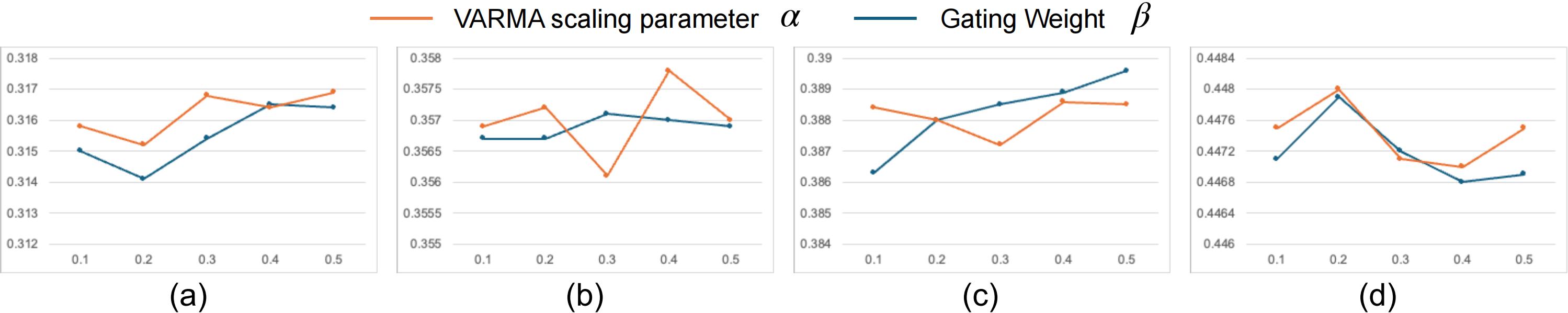}
 	\caption{\textbf{Impact of scaling parameter $\alpha$ and gate $\beta$ on ETTm1.} }
 	\label{hyper}
\end{figure}

We analyze the impact of the VARMA scaling parameter $\alpha$ and the Gating Weight $\beta$. We vary each parameter from 0.1 to 0.5 while keeping the other fixed and evaluated the model's performance on the ETTm1 dataset across four prediction horizons. The results are visualized in Fig.~\ref{hyper}, where subplots (a), (b), (c), and (d) correspond to prediction lengths of 96, 192, 336, and 720, respectively. The primary observation from this study is that our model demonstrates remarkable robustness to the specific settings of both $\alpha$ and $\beta$. As shown across all subplots, the MSE exhibit only marginal fluctuations as the parameters change. The variations in performance are consistently minor, typically occurring at the third or fourth decimal place. This indicates that while our proposed mechanisms are crucial to the model's overall superiority, the precise weighting of their contributions is not a critical factor, and the model is not prone to overfitting these specific hyperparameters. We hypothesize that this stability stems from the synergistic nature of our architecture. The model learns to leverage both the local statistical features from the VFE and the global dependencies from the VE-atten in a balanced manner. This robustness simplifies the tuning process and enhances the model's practical applicability, as it does not require extensive, dataset-specific hyperparameter optimization to achieve strong results.

\subsection{Visualization}

\begin{figure}[t]
 	\centering
 	\includegraphics[width=12cm]{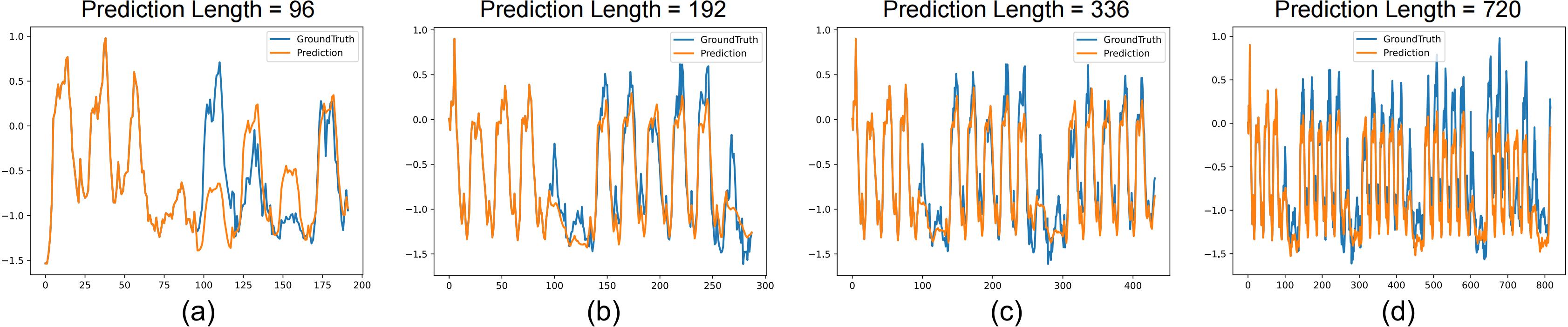}
 	\caption{\textbf{Visualization of forecasting results on Electricity dataset.} }
 	\label{visual}
\end{figure}
Fig.~\ref{visual} presents a qualitative visualization of our model's forecasting performance on the Electricity dataset for varying prediction lengths: (a) 96, (b) 192, (c) 336, and (d) 720 steps. Each subplot displays a randomly selected sample from the test set, comparing the model's prediction (orange) against the ground truth (blue). The visualizations clearly demonstrate our model's strong capability in capturing the complex, multi-scale periodicities inherent in the electricity consumption data. Even in the most challenging long-term forecasting scenario (d, 720 steps), the model accurately predicts both the phase and amplitude of the dominant daily and weekly patterns. The predicted series closely tracks the ground truth, successfully reproducing the sharp peaks and troughs characteristic of load data. This qualitative evidence reinforces the quantitative results, confirming our model's effectiveness in handling intricate, real-world time series with high precision.

\section{Conclusion}

In this paper, we introduced \textit{VARMAformer}, a novel hybrid architecture that synergistically fuses the principles of classical VARMA models with the power of modern Transformers for time series forecasting. Our core contributions are twofold: a dedicated VFE module that explicitly captures local, patch-level statistical dependencies, and the VE-atten mechanism that dynamically modulates queries based on temporal context. Through extensive experiments on several real-world benchmark datasets, we demonstrate that \textit{VARMAformer} consistently outperforms a wide range of state-of-the-art models, particularly in challenging long-term forecasting scenarios. Ablation studies further validate the individual and synergistic contributions of our proposed components. Our work underscores the significant potential of a principled integration of statistical priors into deep learning architectures, paving the way for more robust, accurate, and interpretable time series forecasting models.

\section{Acknowledgements}

This work was supported by grants from the Natural Science Foundation of China (72473148).
%
%
%
%

\bibliographystyle{splncs04}
\bibliography{mybib}

\end{document}